*Computers, Materials & Continua*
*DOI:10.32604/cmc.202x.xxxxxx*
*Type: xxx*

**Tech Science Press**# A Two-Phase Paradigm for Joint Entity-Relation Extraction

**Bin Ji[1], Hao Xu[1], Jie Yu[1], Shasha Li[1], Jun Ma[1], Yuke Ji[2,*] and Huijun Liu[1]**

[1]College of Computer, National University of Defense Technology, Changsha, 410073, China

[2] The Affiliated Eye Hospital of Nanjing Medical University, Nanjing, 210029, China

[*]Corresponding Author: Yuke Ji. Email: jiyuke93@163.com

Received: 9 March 2022; Accepted: XX Month 202X**Abstract:** An exhaustive study has been conducted to investigate span-based models for the joint entity and relation extraction task. However, these models sample a large number of negative entities and negative relations during the model training, which are essential but result in grossly imbalanced data distributions and in turn cause suboptimal model performance. In order to address the above issues, we propose a two-phase paradigm for the span-based joint entity and relation extraction, which involves classifying the entities and relations in the first phase, and predicting the types of these entities and relations in the second phase. The two-phase paradigm enables our model to significantly reduce the data distribution gap, including the gap between negative entities and other entities, as well as the gap between negative relations and other relations. In addition, we make the first attempt at combining entity type and entity distance as global features, which has proven effective, especially for the relation extraction. Experimental results on several datasets demonstrate that the span-based joint extraction model augmented with the two-phase paradigm and the global features consistently outperforms previous state-of-the-art span-based models for the joint extraction task, establishing a new standard benchmark. Qualitative and quantitative analyses further validate the effectiveness the proposed paradigm and the global features.

**Keywords:** Joint extraction; span-based; named entity recognition; relation extraction; data distribution; global features## 1 Introduction

Span-based joint entity and relation extraction models simultaneously conduct NER (**N**amed **E**ntity **R**ecognition) and RE (**R**elation **E**xtraction) in text span forms. Typically, these models are constructed as follows: given an unstructured text, the model divides it into text spans; it then constructs ordered span pairs (a.k.a. relation tuples); and finally, it obtains entities and relations by performing classifications on the semantic representations of spans and relation tuples, respectively. We present a typical case study in Fig. 1: the "In", "In 1831", and "James Garfield" are three span examples; the <"James Garfield", "U.S."> and <"James Garfield", "Ohio"> are two relation tuple examples; a span-based model predicts the types of spans and relation tuples by performing classifications on related semantic representations. For instance, the "In" is classed as the Not-Entity type, and the <"James Garfield", "Ohio"> is classified as the Live type.[1]

---

[1] Span-based models add a Not-Entity type for spans that are not entities and a Not-Relation type for relation tuples that don't hold relations.

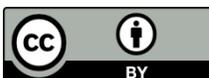

This work is licensed under a Creative Commons Attribution 4.0 International License, which permits unrestricted use, distribution, and reproduction in any medium, provided the original work is properly cited.

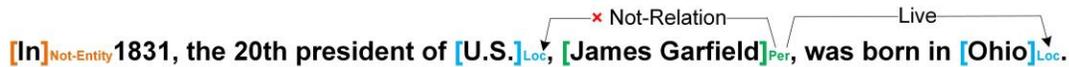

**Figure 1:** An example of the span-based joint extraction. The Loc and Per are two pre-defined entity types, and the Live is a pre-defined relation type. × denotes the <"James Garfield", "U.S.">, supposedly classified into the Live type, is actually classified into the Not-Relation type by SpERT [1], a **Sp**an-based joint **E**ntity-**R**elation extraction model with **T**ransformer.

Span-based joint extraction models [2-7] sample numerous negative entities and relations (i.e., spans of the Not-Entity type and relation tuples of the Not-Relation type) during the model training. These negative examples actually lead to grossly imbalanced data distributions, which is one of the primary reasons for the suboptimal model performance. As shown in Tab. 1, the entity distribution between Other and Not-Entity is 592: 101555 (approximate to 1: 172), the relation distribution between Kill and Not-Relation is 229: 12915 (approximate to 1: 56). Paradoxically, previous work [1] demonstrates that an adequate number of negative examples are required to ensure that the model performs well. Thus, resolving the issue of grossly imbalanced data distributions while maintaining an adequate number of negative examples is a feasible way to improve the model performance.

**Table 1:** The entity and relation counts in CoNLL04, which are obtained using SpERT. The Other, Org, Per and Loc are four pre-defined entity types, and the Kill, LocIn, Work, OrgBI and Live are five pre-defined relation types, and the Not-Entity and Not-Relation are the types of negative entities and relations, respectively.

| NER | Other | Org | Per | Loc | Not-Entity | |
|---|---|---|---|---|---|---|
| | 592 | 786 | 1,370 | 1,541 | 101,555 | |
| RE | Kill | LocIn | Work | OrgBI | Live | Not-Relation |
| | 229 | 312 | 325 | 317 | 421 | 12,915 |

Global features, such as those derived from entity information, can be critical in the joint extraction task. As illustrated in Fig. 1, if SpERT is aware that the "James Garfield" is a person (Per) entity and the "U.S." is a location (Loc) entity beforehand, it may easily classify the <"James Garfield", "U.S."> into the Live type. Moreover, entity distance, which tracks the word counts between two entities, can reflect the entities' correlation. For example, in the CoNLL04 dataset, relations with an entity distance of less than 6 account for 64.5%, and the smaller the distance, the more likely the two entities have a relation. However, as far as we know, previous work [8-12] has used either entity type or entity distance but not both. The combination of the above two types of information may play a more important role in the joint extraction task. As shown in Tab. 2, the <Loc, Loc> tends to have the LocIn relation when the entity distance is smaller, such as 76.6% for [0-3], 12.8% for [4-7] and 3.5% for [8-11], whereas the <Per, Per> tends to have the Kill relation in the case of a bigger entity distance, such as 21.3% for [0-3], 33.5% for [4-7], and 26.7% for [8-11].

In this paper, we propose a two-phase span-based model for the joint extraction task, with the goal of addressing the issue of grossly imbalanced data distributions and the lack of effective global features. Motivated by the fact that we can achieve NER (RE) in two steps, namely first classify all entities (relations) and then predict their types. We divide the joint extraction task into two phases, with the first phase obtaining entities and relations and the second phase predicting their types. Our model reduces the data distribution gap by dozens of times using the two-phase paradigm. Take the data in Tab. 1 as an example: (1) in the first stage, the entity distribution can be reduced to 1: 24 and the relation distribution to 1: 8, whereas the corresponding values in SpERT are 1: 172 and 1: 56, respectively.[2] (2) In the second phase, our model predicts the types of entities and relations, implying that the data distributions are roughly even.[3] Moreover, we attempt for the first time to combine entity type and entity distance as global features and use them to augment our model. Furthermore, we propose a gated mechanism for fusing various semantic representations, taking the weighted importance of each representation into account. In Section 4.5, we validate the effectiveness of the above model components.

---
[2] 1: 24 ≈ (592+786+1370+1541): 101555; 1: 8 ≈ (229+312+325+347+421): 12915.

**Table 2:** Entity distance statistics of the CoNLL04 dataset. We use the ordered entity type tuple to denote ordered relation tuples of the same type, such as the <Per, Loc> denotes all relation tuples that the type of their first entity is the Per and the type of their second entity is the Loc. We divide all distances into four distance intervals, i.e., [0-3], [4-7], [8-11] and [>11].

| Relation tuple | Type | Entity distance | | | |
|---|---|---|---|---|---|
| | | [0-3] | [4-7] | [8-11] | [>11] |
| <Per, Loc> | Live | 39.0% | 27.6% | 10.5% | 22.9% |
| <Loc, Loc> | LocIn | 76.6% | 12.8% | 3.5% | 7.1% |
| <Per, Org> | Work | 41.5% | 35.1% | 8.0% | 15.4% |
| <Org, Loc> | OrgBI | 72.0% | 10.7% | 5.5% | 11.8% |
| <Per, Per> | Kill | 21.3% | 33.5% | 26.7% | 18.5% |

Experimental results on the ACE05, CoNLL04 and SciERC datasets demonstrate that our model consistently outperforms the strongest span-based baselines in terms of F1-score, providing a new span-based benchmark for the joint extraction task. Extensive analyses further validate the effectiveness of our model.

In summary, our model differs from the previous span-based models in three ways: (1) As far as we know, our model makes the first attempt to balance the grossly imbalanced data distributions. (2) Our model combines entity type and entity distance as the global features, whereas previous span-based models use at most one of them. (3) Our model uses a gated mechanism to fuse various semantic representations, whereas previous span-based models use a simple concatenation manner.

**2 Related Work**

*2.1 Span-based Joint Entity and Relation Extraction*

Recently, span-based models have been extensively investigated for the joint entity and relation extraction task. Luan et al. [2] propose almost the first span-based joint model and attempt to further improve model performance by incorporating the coreference resolution task [13-14]. Luan et al. [4] also include the coreference resolution task in their span-based joint model. Moreover, some other span-based models [5] have examined how to incorporate additional natural language processing tasks, such as event detection [15-16]. More recently, Dixit and Al-Onaizan [3] introduce the pre-trained language model, i.e., ELMo (**E**mbeddings from **L**anguage **Mo**dels) [17], into the span-based joint model for the first time. Eberts and Ulges [1] propose to use BERT (**B**idirectional **E**ncoder **R**epresentation from **T**ransformers) [18] as the backbone of their span-based joint model. Zhong and Chen [7] propose to use ALBERT (**A L**ite **BERT**) [19] in their span-based joint model. However, these models suffer from grossly imbalanced data distributions, as the span-based paradigm requires extensive negative entities and relations. Although our model also samples a large number of examples, we propose a two-phase paradigm to eliminate the data distribution gap effectively.

---

[3] The entity distributions can be approximate to (592: 786: 1370: 1541), and the relation distributions can be approximate to (229: 312: 325: 347: 421), which are approximately even.

*2.2 Global Features*

The entity type and entity distance are two types of important global features that are frequently used in joint extraction models [20-27]. Miwa and Bansal [18], Sun and Grishman [28], and Bekoulis et al. [9], are among the first to use entity types as global features in their joint extraction models. They concatenate fixed-size embeddings trained for entity types to relation semantic representations. Zhao et al. [10] model strong correlations between entity labels and text tokens and concatenate entity label embeddings to relation semantic representations. For entity distance, Zeng et al. [11] and Ye et al. [12] concatenate relative entity position features to relation semantic representations. However, the above models use either entity type or entity distance but make no attempt to combine them. In comparison, our model suggests combining the entity type and entity distance as global features, which is validated to be more effective.

**3 Model**

The neural architecture of our two-phase span-based model is illustrated in Fig. 2. For a given unstructured text $\mathcal{T} = (t_1, t_2, \ldots, t_n)$ where $t_i$ denotes the $i$-$th$ text token, our model first obtains its BERT embedding sequence (Section 3.1); then in Phase One, our model obtain entities and relations by performing binary classifications on semantic representations of spans and relation tuples, respectively. These entities and relations are referred to as coarse-grained entities and relations, respectively (Section 3.2); next, in Phase Two, our model predicts the types of these coarse-grained entities and relations, obtaining fine-grained entities and relations (Section 3.3). In both phases, we combine the entity type and entity distance as global features and use a gated mechanism to fuse various semantic representations.

We formulate the text spans (denoted as $\mathcal{S}$) from $\mathcal{T}$ as follows:

$$\mathcal{S} = (t_j, t_{j+1}, \ldots, t_{j+k}) \quad s.t. \quad 0 \leq j \leq j+k \leq n \text{ and } k \leq \epsilon, \tag{1}$$

where $\epsilon$ is the span width threshold.

*3.1 Embedding Layer*

Our model uses the BERT [18] model as the word embedding generator. We denote the BERT embedding sequence for text $\mathcal{T}$ as follows:

$$\boldsymbol{E}_{\mathcal{T}} = (\boldsymbol{x}_0, \boldsymbol{x}_1, \boldsymbol{x}_2, \ldots, \boldsymbol{x}_n), \tag{2}$$

where $\boldsymbol{E}_{\mathcal{T}} \in \mathbb{R}^{(n+1)*d}$ and $d$ is the BERT embedding dimension. $\boldsymbol{x}_0$ is the BERT embedding for the added [CLS] token, which is a built-in setting of the BERT model.[4] $\boldsymbol{x}_i$ is the BERT embedding for the token $t_i$. Due to the fact that BERT may tokenize a token into several sub-tokens in order to avoid the Out-of-Vocabulary (OOV) problem, we obtain $\boldsymbol{x}_i$ by applying the max-pooling function to the BERT embeddings of the sub-tokens tokenized from the token $t_i$.

Based on $\boldsymbol{E}_{\mathcal{T}}$, We denote the BERT embedding sequence for span $\mathcal{S}$ as follows:

$$\boldsymbol{E}_{\mathcal{S}} = (\boldsymbol{x}_j, \boldsymbol{x}_{j+1,}, \ldots, \boldsymbol{x}_{j+k}). \tag{3}$$

*3.2 Phase One*

As shown in Fig. 2, the Phase One is composed of two modules: **Entity Classification** and **Relation Classification**, where the former obtains coarse-grained entities and the latter obtains coarse-grained relations.

*3.2.1 Entity Classification*

---

[4] The [CLS] token is a specific token that added to the beginning of tokenized texts. The embedding of the [CLS] token is generally used for text classifications.

This module obtains coarse-grained entities by performing binary classification on span semantic representations. We begin by converting all entity types in the training set to the Entity type and set the type of sampled negative entities to the Not-Entity type. Our model will be trained to classify spans as the Entity type when they are predicted to be entities, otherwise the Not-Entity type.

In this paper, we obtain the span semantic representations using three different types of semantic representations: (1) span token representation, (2) contextual representation, and (3) span width embedding.

For the span $\mathcal{S}$, we obtain its token representation (denoted as $\widehat{E}_\mathcal{S}$) by applying the max-pooling function to its BERT embedding sequence $E_\mathcal{S}$:

$$\widehat{E}_\mathcal{S} = [\max(x_{j,1}, x_{j+1,1}, \ldots, x_{j+k,1}), \max(x_{j,2}, x_{j+1,2}, \ldots, x_{j+k,2}), \ldots, \max(x_{j,d}, x_{j+1,d}, \ldots, x_{j+k,d})], \quad (4)$$

where $\widehat{E}_\mathcal{S} \in \mathbb{R}^d$.

In this paper, we take the $x_0 \in \mathbb{R}^d$ as the contextual representation for any span $\mathcal{S}$ from the text $\mathcal{T}$.

Span width embedding allows the model to incorporate prior experience over span widths. In this paper, we train a fixed-size embedding for each span width (i.e., 1, 2 ,...) during the model training. And we refer to the width embedding for the span $\mathcal{S}$ (length is $j + 1$) as $W_{j+1}$, where $W_{j+1} \in \mathbb{R}^d$.

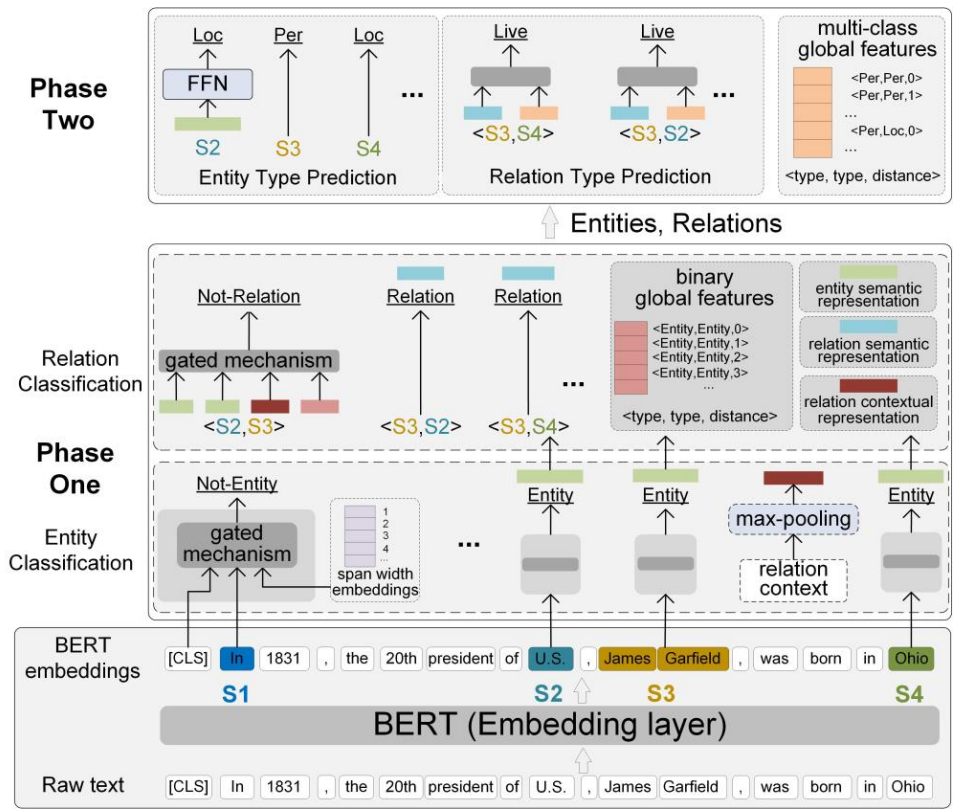

**Figure 2:** Neural architecture of the proposed model. In the Phase One, the model classifies entities and relations; in the Phase Two, the model predicts their types. In both phases, the model combines entity type and distance as global features.

$\widehat{E}_\mathcal{S}$, on the other hand, should theoretically contribute the most to the span semantic representation, whereas $W_{j+1}$ the least. However, the previous work [1,4] has overlooked this critical property, concatenating the above representations, which has been demonstrated to be insufficient [10]. In this paper, we propose a gated mechanism that enables us to weigh the importance of each representation. The span semantic representation (denoted as $\boldsymbol{E}'_S$) is then obtained by summing the weighted representations:

$$v_o = \boldsymbol{W}_1 \boldsymbol{E}_o + b_1 \quad s.t. \quad 1 \leq o \leq \delta, \tag{5a}$$

$$\alpha_o = \frac{\exp^{v_o}}{\sum_{m=1}^{\delta} \exp^{v_m}}, \tag{5b}$$

$$\boldsymbol{E}'_S = \sum_{m=1}^{\delta} \alpha_m \boldsymbol{E}_m, \tag{5c}$$

where $\boldsymbol{W}_1 \in \mathbb{R}^d$, $b_1$ is a scalar and $\{\boldsymbol{E}_o, \boldsymbol{E}'_S\} \in \mathbb{R}^d$. In the current scenario, the $\delta$ is set to 3, and $\boldsymbol{E}_1$, $\boldsymbol{E}_2$, and $\boldsymbol{E}_3$ are $\widehat{\boldsymbol{E}}_S$, $\boldsymbol{x}_0$, and $\boldsymbol{W}_{j+1}$, respectively.

To obtain coarse-grained entities, we first pass the $\boldsymbol{E}'_S$ through an FFN (**F**eed **F**orward **N**etwork), and then feed it into the sigmoid function, which yields probability distributions for the span $S$ on the above two types, i.e., Entity and Not-Entity:

$$\widehat{\boldsymbol{E}}'_S = \boldsymbol{W}_2 \boldsymbol{E}'_S + \boldsymbol{b}_2, \tag{6a}$$

$$\boldsymbol{y}_{S,i} = \frac{1}{1 + \exp^{\widehat{\boldsymbol{E}}'_{S,i}}}, \tag{6b}$$

where $\boldsymbol{W}_2 \in \mathbb{R}^{2*d}$ and $\boldsymbol{b}_2 \in \mathbb{R}^2$ are trainable FFN parameters, $\widehat{\boldsymbol{E}}'_S \in \mathbb{R}^2$. By searching the highest-scored class, $\boldsymbol{y}_S$ estimates whether $S$ is a coarse-grained entity or not. We build a coarse-grained entity set $T_e$ with the predicted entities.

*3.2.2 Relation Classification*

This module obtains coarse-grained relations by performing binary classification on semantic representations of relation tuples. We begin by converting all relation types in the training set to the Relation type and assigning the Not-Relation type to sampled negative relations. Our model will be trained to classify relation tuples as the Relation type if they have relations, otherwise the Not-Relation type.

Let $e_2$ and $e_2$ be any two coarse-grained entities of $T_e$. We formulate relation tuples as follows:

$$r_b = <e_1, e_2> \quad s.t. \quad e_1, e_2 \in T_e \text{ and } e_1 \neq e_2. \tag{7}$$

We obtain the sematic representation of $r_b$ with four different types of semantic representations, namely (1) the representation of $e_1$, (2) the representation of $e_2$, (3) relation contextual representation, and (4) global features. We use $\boldsymbol{E}'_{e_1}$ and $\boldsymbol{E}'_{e_2}$, which are calculated using the Eq. (5) in Section 3.2.1, as the representations of $e_1$ and $e_2$, respectively.

Relation context is the text that between the two entities of a relation tuple [29]. In this paper, we assume the relation context of $r_b$ as $Con = (\boldsymbol{t}_p, \boldsymbol{t}_{p+1}, \dots, \boldsymbol{t}_{p+q})$. Thus the BERT embedding sequence for $Con$ is as follows:

$$\boldsymbol{E}_c = (\boldsymbol{x}_p, \boldsymbol{x}_{p+1}, \dots, \boldsymbol{x}_{p+q}). \tag{8}$$

We obtain the contextual representation of $r_b$ (denoted as $\boldsymbol{R}_c$) by applying the max-pooling function to $\boldsymbol{E}_c$:

$$\boldsymbol{R}_C = [\max(\boldsymbol{x}_{p,1}, \boldsymbol{x}_{p+1,1}, \dots, \boldsymbol{x}_{p+q,1}), \max(\boldsymbol{x}_{p,2}, \boldsymbol{x}_{p+1,2}, \dots, \boldsymbol{x}_{p+q,2}), \dots, \max(\boldsymbol{x}_{p,d}, \boldsymbol{x}_{p+1,d}, \dots, \boldsymbol{x}_{p+q,d})]. \tag{9}$$

In this paper, we propose to combine entity type and entity distance as global features. Due to the fact that all entities here are the Entity type, only the entity distance can be used to distinguish different feature entries. As show in Fig. 2, we refer to them as binary global features. During model training, we train a fixed-size embedding for each feature entry and denote the feature embedding for $r_b$ as $\boldsymbol{D}_{r_b}$, where $\boldsymbol{D}_{r_b} \in \mathbb{R}^d$.

We obtain the semantic representation of $r_b$ (denoted as $\boldsymbol{R}_{r_b}$) using the proposed gated mechanism, as shown in Eq. (5). In the current scenario, the $\delta$ is set to 4, and $\boldsymbol{E}_1$, $\boldsymbol{E}_2$, $\boldsymbol{E}_3$, and $\boldsymbol{E}_4$ are $\boldsymbol{E}'_{e_1}$, $\boldsymbol{E}'_{e_2}$, $\boldsymbol{R}_C$, and $\boldsymbol{D}_{r_b}$, respectively.

To obtain coarse-grained relations, we first pass the $\boldsymbol{R}_{r_b}$ through an FFN and then feed it into the sigmoid function, which yields probability distributions for $r_b$ on the above two types, i.e., Relation and Not-Relation:

$$\widehat{\boldsymbol{R}}_{r_b} = \boldsymbol{W}_3 \boldsymbol{R}_{r_b} + \boldsymbol{b}_3, \tag{10a}$$

$$\boldsymbol{y}_{r_b,i} = \frac{1}{1+\exp^{\widehat{R}_{r_b,i}}}, \tag{10b}$$

where $\boldsymbol{W}_3 \in \mathbb{R}^{2*d}$ and $\boldsymbol{b}_3 \in \mathbb{R}^2$ are trainable FFN parameters, and $\widehat{\boldsymbol{R}}_{r_b} \in \mathbb{R}^2$. By searching the highest-scored class, $\boldsymbol{y}_{r_b}$ estimates whether $r_b$ has a relation or not. We build a coarse-grained relation set $T_r$ with the predicted relations.

### 3.2.3 Training Loss of Phase One

For each of the above two binary classifications, the training objective is to minimize the following binary cross-entropy loss:

$$\mathcal{L}_b^t = -\frac{1}{N^t}\sum_{i=1}^{N^t}(\boldsymbol{y}_i^t \log \widehat{\boldsymbol{y}}_i^t + (1-\boldsymbol{y}_i^t)(1-\widehat{\boldsymbol{y}}_i^t)), \tag{11}$$

where $t$ denotes one of the above two classifications. $\boldsymbol{y}_i^t$ is the one-hot vector of gold type. $\widehat{\boldsymbol{y}}_i^t$ is the predicted probability distributions. $N^t$ is the number of instances for the classification $t$.

### 3.3 Phase Two

In the Phase Two, our model predicts the types of coarse-grained entities and relations, obtaining fine-grained entities and relations. The Phase Two, as illustrated in Fig. 2, is composed of two modules: **Entity Type Predication** and **Relation Type Predication**.

### 3.3.1 Entity Type Predication

In this module, we obtain entity types by conducting multi-class classifications on the semantic representations of coarse grained entities. Specifically, for each coarse-grained entity $e$ in $T_e$, we denote its semantic representation as $\boldsymbol{E}'_e \in \mathbb{R}^d$, which is obtained the same as the span semantic representation, as illustrated in Section 3.2.1. To obtain the type of $e$, we first pass $\boldsymbol{E}'_e$ through an FFN and then feed it into the softmax function, which yields probability distributions for $e$ on $\Omega$, where $\Omega$ is the set of all pre-defined entity types:

$$\widehat{\boldsymbol{E}}'_e = \boldsymbol{W}_4 \boldsymbol{E}'_e + \boldsymbol{b}_4, \tag{12a}$$

$$\widehat{\boldsymbol{y}}_{e,i} = \frac{1}{\sum_{j=1}^{|\Omega|} \exp^{\widehat{E}'_{e,j}}} \quad s.t. \ \ 1 \leq i \leq |\Omega|, \tag{12b}$$

where $\boldsymbol{W}_4 \in \mathbb{R}^{|\Omega|*d}$ and $\boldsymbol{b}_4 \in \mathbb{R}^{|\Omega|}$ are trainable FFN parameters. $|\Omega|$ is the counts of pre-defined entity types. By searching the highest-scored class, $\widehat{\boldsymbol{y}}_e$ estimates a pre-defined entity type for $e$.

### 3.3.2 Relation Type Predication

We obtain relation types by performing multi-class classifications on relation semantic representations. As shown in Fig. 2, the relation semantic representation is derived from two parts: the relation representation used for the binary relation classification and multi-class global features.

For each coarse-grained relation $r$ in $T_r$, we denote its representation used for the binary relation classification as $\boldsymbol{R}_r$, which can be obtained using the same approach illustrated in Section 3.2.2. As shown in Fig. 2, we combine the entity type and entity distance as the multi-class global features. We formulate the combination of entity type and entity distance as follows:

$$C = \{\Omega \otimes \Omega \otimes \Delta\}, \tag{13}$$

where $\Omega$ is the set of pre-defined entity types, $\Delta$ is the set of entity distances, and $\otimes$ denotes the Cartesian Product. For each feature entry in $C$, we train a fixed-size embedding for it during the model training. We denote the feature embedding for $r$ as $\boldsymbol{C}_r \in \mathbb{R}^d$.

Then we obtain the relation semantic representation (denoted as $\boldsymbol{R}'_r$) with $\boldsymbol{R}_r$ and $\boldsymbol{C}_r$, which is calculated using the Eq. (5). In the current scenario, the $\delta$ is set to 2, and $\boldsymbol{E}_1$ and $\boldsymbol{E}_2$ are $\boldsymbol{R}_r$ and $\boldsymbol{C}_r$, respectively.

To obtain the type of $r$, we first pass $\boldsymbol{R}'_r$ through an FFN, and then feed it into the softmax function, which yields probability distributions for $r$ on $\Psi$, where $\Psi$ is the set of all pre-defined relation types:

$$\widehat{\boldsymbol{R}}'_r = \boldsymbol{W}_5 \boldsymbol{R}'_r + \boldsymbol{b}_5, \tag{14a}$$

$$\widehat{\boldsymbol{y}}_{r,i} = \frac{\exp^{\widehat{R}'_{r,i}}}{\sum_{j=1}^{|\Psi|} \exp^{\widehat{R}'_{r,j}}} \quad s.t. \quad 1 \leq j \leq |\Psi|, \tag{14b}$$

where $\boldsymbol{W}_5 \in \mathbb{R}^{|\Psi|*d}$ and $\boldsymbol{b}_5 \in \mathbb{R}^{|\Psi|}$. $|\Psi|$ is the counts of pre-defined relation types. By searching the highest-scored class, $\widehat{\boldsymbol{y}}_r$ estimates the type that $r$ has.

*3.3.3 Training Loss of the Phase Two*

For each of above two multi-class classification tasks, the training objective is to minimize the following cross-entropy loss:

$$\mathcal{L}_p^{t'} = -\frac{1}{M^{t'}} \sum_{i=1}^{M^{t'}} \boldsymbol{y}_i^{t'} \log \widehat{\boldsymbol{y}}_i^{t'}, \tag{15}$$

where $t'$ denotes one of the above two classifications. $\boldsymbol{y}_i^{t'}$ is the one-hot vector of gold type. $\widehat{\boldsymbol{y}}_i^{t'}$ is the predicted probability distributions. $M^{t'}$ is the number of instances for the classification $t'$.

*3.4 Model Training*

During the model training, we minimize the following joint training loss:

$$\mathcal{L}(\boldsymbol{W}; \boldsymbol{\theta}) = \sum_{t \in T} \mathcal{L}_b^t + \sum_{t' \in T'} \mathcal{L}_p^{t'}, \tag{16}$$

where $T$ denotes the two binary classifications and $T'$ denotes the two multi-class classifications.

**4 Experiments**

*4.1 Dataset*

We evaluate our model on ACE05 [30], CoNLL04 [31], and SciERC [2].

ACE05 defines seven entity types (Per, Org, Loc, Gpe, Fac, Veh, and Wea) and six relation types (Phys, Part-whole, Per-soc, Org-aff, Art, and Gen-aff) between entities. We use the same data splits, pre-processing, and task settings proposed by Li and Ji [32] and Li et al. [33]. It has 351 documents for training, 80 for development and 80 for test.

CoNLL04 defines four entity types (Loc, Org, Per, and Other) and five relation types (Kill, Live, LocIn, OrgBI, and Work). We use the splits defined by Ji et al. [6] and Wang et al. [25]. The dataset consists of 910 instances for training, 243 for development and 288 for test.

SciERC is derived from 500 abstracts of AI papers. The dataset defines six scientific entities (Task, Method, Metric, Material, Other, and Generic) and seven relation types (Compare, Conjunction, Evaluate-for, Used-for, Feature-of, Part-of, and Hyponym-of) in a total of 2,687 sentences. We use the same training (1,861 sentences), development (275 sentences), and test (551 sentences) split following the previous work [3,34].

*4.2 Experimental Setup*

For a fair comparison with previous work, we use the bert-base-cased model on ACE05 and CoNLL04, and use the scibert-scivocab-cased model on SciERC. We optimize our model using the BertAdam for 120 epochs with a learning rate of 5e-5 and a weight decay of 1e-2. We set the span width threshold $\epsilon$ to 10 for all datasets and the entity distance set $\Delta$ to $\{0, 1, ..., 10\}$, and if an entity distance is greater than 10, we set it to 10. Moreover, we employ the same negative sampling strategy proposed by Eberts and Ulges [1]. We use the standard Precision (P), Recall (R) and F1-score to evaluate the model performance:

$$P = \frac{TP}{TP+FP}, \tag{17a}$$

$$R = \frac{TP}{TP+FN}, \tag{17b}$$

$$F1 = \frac{2*P*R}{P+R}, \tag{17c}$$

where TP, FP and FN stand for true positive, false positive, and false negative, respectively.

For ACE05, an entity mention is considered correct if its head region and type match the ground truth, and a relation is correct if both its relation type and two entity mentions are correct. For CoNLL04, an entity mention is considered correct if its offsets and type match the ground truth, and a relation is correct if both its relation type and two entity mentions are correct. For SciERC, the entity type is not considered when evaluating relation extraction, which is in line with the previous work [6,7]. And the remaining settings are identical to those for CoNLL04.

### 4.3 Main Results

We compare our model with all the published span-based models for the joint extraction task that we are aware of. We report the comparison results in Tab. 3, Tab. 4, and Tab. 5, from which we can observe that our model consistently outperforms the strongest baselines in terms of F1-score across the three datasets.

To be more precise, on ACE05, our model achieves +0.4% and +3.2% absolute F1 gains on NER and RE, respectively, when compared to Ji et al. [6] that achieves the previous best NER performance. In addition, when compared to Zhong and Chen [7] that achieves the previous best RE performance, our model achieves +1.3% and +1.4% absolute F1 gains on NER and RE, respectively. On CoNLL04, our model achieves +0.3% and +1.6% absolute F1 gains on NER and RE, respectively, when compared to the strongest baseline Ji et al. [6]. On SciERC, when compared to Santosh et al. [35] that achieves the previous best NER performance, our model delivers +0.5% and +1.4% absolute F1 gains. When compared to Zhang et al. [36] that achieves the previous best RE results, our model achieves +0.6% and +0.2% absolute F1 gains.

**Table 3:** Performance comparisons on ACE05. * denotes using the bert-based-cased model and the single-sentence setting for a fair comparison. The bold values denote the best results.

| Model | NER | | | RE | | |
|---|---|---|---|---|---|---|
| | P | R | F1 | P | R | F1 |
| Dixit and Al-Onaizan [3] | 85.9 | 86.1 | 86.0 | 68.0 | 58.4 | 62.8 |
| Luan et al. [4] | - | - | 88.4 | - | - | 63.2 |
| Wadden et al. [5] | - | - | 88.6 | - | - | 63.4 |
| Zhong and Chen [7] * | - | - | 88.7 | - | - | 66.7 |
| Wang et al. [37] | - | - | 88.9 | - | - | 64.3 |
| Ji et al. [6] | 89.3 | 89.9 | 89.6 | 71.2 | 60.2 | 65.2 |
| Our Model | **89.4** | **90.6** | **90.0** | **72.8** | **64.5** | **68.4** |

We attribute the above performance improvements to that our model is capable of balancing the grossly imbalanced data distributions and exploiting the effective global features.

**Table 4:** Performance comparisons on CoNLL04. The bold values denote the best results.

| Model | NER | | | RE | | |
|---|---|---|---|---|---|---|
| | P | R | F1 | P | R | F1 |
| Eberts and Ulges [1] | 88.3 | 89.6 | 89.0 | 73.0 | 70.0 | 71.5 |
| Zhang et al. [36] | 88.1 | 90.6 | 89.3 | 75.4 | 71.1 | 73.2 |
| Tang et al. [38] | - | - | 89.4 | - | - | 72.6 |
| Ji et al. [6] | **90.1** | 90.4 | 90.2 | 77.0 | 71.9 | 74.3 |
| Our Model | 89.6 | **91.4** | **90.5** | **78.7** | **73.3** | **75.9** |

**Table 5:** Performance comparisons on SciERC. ‡ denotes using the scibert-scivocab-cased model. The bold values denote the best results.

| Model | NER | | | RE | | |
|---|---|---|---|---|---|---|
| | P | R | F1 | P | R | F1 |
| Luan et al. [2] | 67.2 | 61.5 | 64.2 | 47.6 | 33.5 | 39.3 |
| Luan et al. [4] | - | - | 65.2 | - | - | 41.6 |
| Wadden et al. [5] | - | - | 67.5 | - | - | 48.4 |
| Zhong and Chen [7] ‡ | - | - | 68.9 | - | - | 50.1 |
| Eberts and Ulges [1] ‡ | 70.9 | 69.8 | 70.3 | 53.4 | 48.5 | 50.8 |
| Zhang et al. [36] ‡ | 69.7 | 71.1 | 70.4 | **55.3** | 50.0 | 52.5 |
| Santosh et al. [35] ‡ | **69.8** | 71.3 | 70.5 | 51.9 | 50.6 | 51.3 |
| Our Model ‡ | 69.7 | **72.3** | **71.0** | 52.9 | **52.5** | **52.7** |

*4.4 Effectiveness Investigations*

We conduct extensive effectiveness investigations across the three datasets and use SpERT [1] as the baseline. SpERT is the most similar model to ours, and it uses two linear decoders for entity and relation classifications, as well as the BERT model as a backbone. SpERT, on the other hand, ignores the global features and does not balance the imbalanced data distributions. Furthermore, to make a fair comparison, our model employs the same negative sampling strategy as SpERT.

*4.4.1 Data Distributions*

As illustrated in Tab. 6, we compare our model with the baseline in terms of the most imbalanced data distributions. We obtain the data distributions on NER and RE by comparing the numbers of different types of entities and relations, i.e., the smallest number v.s. the largest number. And we obtain the data distributions during the model training. We have the following observations: (1) On ACE05, the most imbalanced data distributions of the baseline are 1: 773.3 on NER and 1: 150.0 on RE. Our model, on the other hand, reduces the ratios to 1: 21.3 and 1: 13.8, respectively. (2) On CoNLL04, the most imbalanced data distributions of the baseline are 1: 171.5 on NER and 1: 56.4 on RE. Our model, on the other hand, reduces the ratios to 1: 23.7 and 1: 9.9, respectively. (3) On SciERC, the most imbalanced data distributions of the baseline are 1: 605.3 on NER and 1: 913.5 on RE. Our model, on the other hand, reduces the ratios to 1: 25.5 and 1: 35.6, respectively.

Based on the above observations, we conclude that the two-phase paradigm allows our model to avoid suffering from grossly imbalanced data distributions.

**Table 6:** Comparisons regarding the most imbalanced data distributions between our model and the baseline.

| Data | Task | Baseline | Our model | |
|---|---|---|---|---|
| | | | Phase One | Phase Two |
| ACE05 | NER | 1: 773.3 | 1: 19.7 | 1: 21.3 |
| | RE | 1: 150.0 | 1: 13.8 | 1:  3.4 |
| CoNLL04 | NER | 1: 171.5 | 1: 23.7 | 1:  2.6 |
| | RE | 1: 56.4 | 1:  9.9 | 1:  1.8 |
| SciERC | NER | 1: 605.3 | 1: 25.5 | 1:  7.7 |
| | RE | 1: 913.5 | 1: 35.6 | 1: 13.6 |

*4.4.2 Effectiveness against Entity Length*

In general, as the entity lengths increase, it becomes increasingly difficult to recognize the entities. In this section, we conduct investigations on NER performance in relation to entity lengths. We divide all entity lengths, which are restricted by the span width threshold $\epsilon$ ($\epsilon$ is set to 10), into five intervals, i.e., [1-2], [3-4], [5-6], [7-8], and [9-10]. We conduct investigations on the dev sets of the three datasets and report the results in Fig. 3. We can observe that our model consistently outperforms the baseline across all length intervals on the three datasets. Moreover, our model obtains greater F1 gains when the entity length increases. To be more precise, our model achieves the greatest improvement on ACE05 when the entity length is [7-8], and on CoNLL04 and SciERC when the entity length is [9-10], suggesting that our model is more successful in the case of long entity lengths.

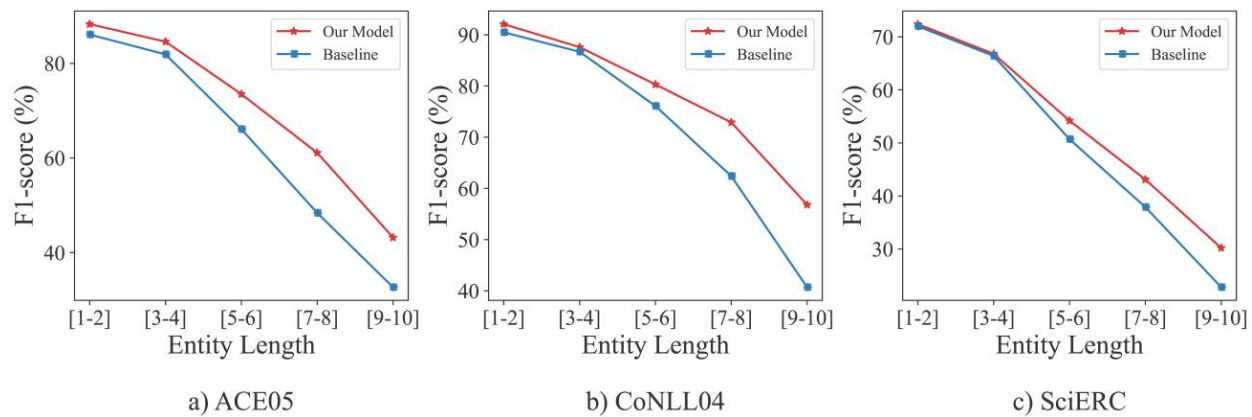

a) ACE05             b) CoNLL04             c) SciERC

**Figure 3:** NER performance (F1-score) comparison of our model and the baseline under various entity length intervals, which are tested on the dev sets of three datasets.

*4.4.3 Effectiveness against Entity Distance*

In general, as the distance between the two entities of a relation increases, the relation becomes more difficult to extract. In this section, we conduct investigations on RE performance in relation to entity distances. We divide all entity distances into five intervals, namely [0], [1-3], [4-6], [7-9], and [>=10]. We conduct investigations on the dev sets of the three datasets and report the investigation results in Fig. 4. The results demonstrate that our model beats the baseline across all distance intervals. Specifically, our model obtains greater improvement when the distance increases, demonstrating that our model is more effective in the case of long entity distances.

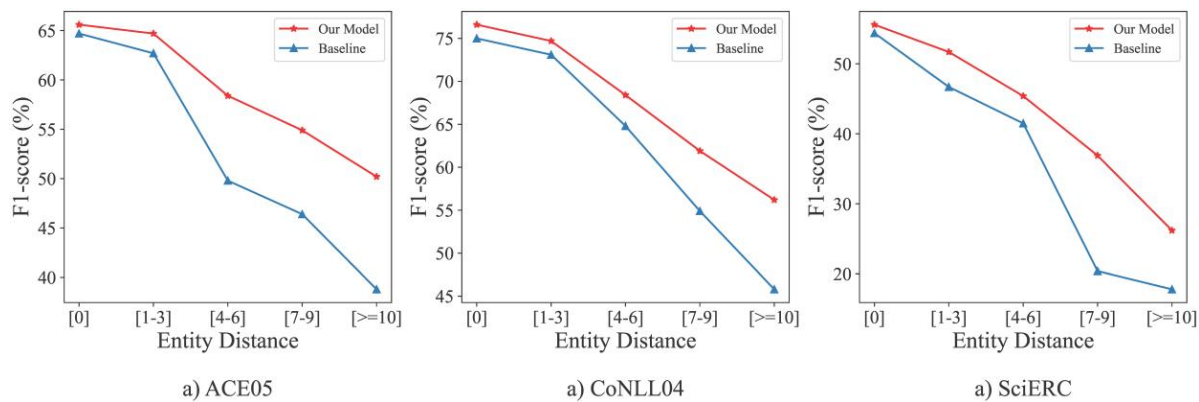

**Figure 4:** RE performance (F1-score) comparison of our model and the baseline under various entity distance intervals, which are tested on the dev sets of three datasets.

*4.5 Ablation Study*

We conduct ablation studies on the dev sets of the three datasets to analyze the effects of various model components. We report the ablation results in Tab. 7, where the "w/o Two-Phase" denotes ablating the two-phase paradigm. As a result, our approach is incapable of dealing with unbalanced data distributions. Additionally, our model cannot make use of binary global features, but retains multi-class global features. The "w/o Bi-Features" denotes ablating the binary global features, which is realized by removing $D_{r_b}$ from $R_{r_b}$. The "w/o Multi-Features" denotes ablating the multi-class global features, which is realized by removing $C_r$ from $R'_r$. The "w/o Both-Features" denotes conducting the above "w/o Bi-Features" and "w/o Multi-Features" ablations simultaneously. The "w/o gated" denotes ablating the gated mechanism. We use the concatenation manner to concatenate various semantic representations instead. The "base" denotes conducting all above ablations. After doing this, our model has the same neural architecture as SpERT.

**Table 7:** Ablation results on the dev sets of the three datasets. We only report the F1-scores.

| Model | ACE05 | | CoNLL04 | | SciERC | |
|---|---|---|---|---|---|---|
| | NER (F1) | RE (F1) | NER (F1) | RE (F1) | NER (F1) | RE (F1) |
| Our Model | 88.1 | 64.8 | 88.7 | 74.2 | 71.4 | 53.5 |
| w/o Two-Phase | 86.5(-0.6) | 61.7(-3.1) | 86.4(-2.3) | 71.5(-2.7) | 68.2(-3.2) | 50.9(-2.6) |
| w/o Bi-Features | 88.3(+0.2) | 63.9(-0.9) | 88.2(-0.5) | 73.5(-0.7) | 71.1(-0.3) | 51.7(-1.8) |
| w/o Multi-Features | 87.8(-0.3) | 63.4(-1.4) | 88.4(-0.3) | 72.4(-1.8) | 71.5(+0.1) | 52.0(-1.5) |
| w/o Both-Features | 87.5(-0.6) | 61.5(-3.3) | 88.5(-0.2) | 72.0(-2.2) | 70.0(-0.4) | 51.5(-2.0) |
| w/o gated | 87.9(-0.2) | 64.3(-0.5) | 88.1(-0.6) | 73.3(-0.9) | 69.9(-1.5) | 52.7(-0.8) |
| base | 86.3(-1.8) | 61.5(-3.3) | 85.9(-2.8) | 70.2(-4.0) | 67.8(-3.6) | 49.0(-4.5) |

We have the following observations: (1) The two-phase paradigm consistently improves the model performance across the three datasets, delivering +0.6% to +3.2% F1-scores on NER and +2.6% to +3.1% F1-scores on RE, which can be attributed to the paradigm's ability to prevent our model from being harmed by grossly imbalanced data distributions. (2) Both binary and multi-class global features consistently benefit RE performance, and the multi-class features are generally more effective than the binary ones, as demonstrated on ACE05 and CoNLL04. The explanation for this could be that the multi-class features take into account fine-grained entity types. Additionally, both types of global features have a negligible effect on NER. A plausible explanation is that these features are derived from entity information and are employed in the relation extraction. (3) The combination of the two types of global features results in improved RE performance, suggesting that they have a beneficial effect on one another. (4) The proposed gated method consistently improves model performance, bringing +0.2% to +1.50% F1-scores on NER and +0.5% to +0.9% on RE, suggesting that the gated mechanism can better fuse various semantic representations.

## 5 Conclusion

In this paper, we propose a two-phase span-based model for the joint entity and relation extraction task, aiming to tackle the grossly imbalanced data distributions caused by the essential negative sampling. And we augment the proposed model with global features obtained by combining entity types and entity distances. Moreover, we propose a gated mechanism for effectively fusing various semantic representations. Experimental results on several datasets demonstrate that our model consistently outperforms the strongest span-based models for the joint extraction task, establishing a new standard benchmark.

**Funding Statement:** This research was supported by the National Key Research and Development Program [2020YFB1006302].

**Conflicts of Interest:** The authors declare that they have no conflicts of interest to report regarding the present study.